\pdfoutput =1

\documentclass[manuscript]{acmart}
\AtBeginDocument{%
  \providecommand\BibTeX{{%
    \normalfont B\kern-0.5em{\scshape i\kern-0.25em b}\kern-0.8em\TeX}}}

\setcopyright{acmcopyright}
\copyrightyear{2021}
\acmYear{2021}
\acmDOI{10.1145/1122445.1122456}

\acmBooktitle{ACM Trans. Intell. Syst. Technol.,
  April, 2021}
\acmPrice{15.00}
\acmISBN{978-1-4503-XXXX-X/18/06}

\usepackage{wrapfig}
\usepackage{subfig}

\usepackage{caption}
\begin{document}

\title{Reinforcement Learning for Quantitative Trading}

\author{Shuo Sun}
\affiliation{%
\institution{Nanyang Technological University}
\country{Singapore}}
\email{shuo.sun@ntu.edu.sg}

\author{Rundong Wang}
\affiliation{%
\institution{Nanyang Technological University}
\country{Singapore}}
\email{rundong001@e.ntu.edu.sg}

\author{Bo An}
\affiliation{%
\institution{Nanyang Technological University}
\country{Singapore}}
\email{boan@ntu.edu.sg}

\renewcommand{\shortauthors}{Sun et al.}


\begin{abstract}
Quantitative trading (QT), which refers to the usage of mathematical models and data-driven techniques in analyzing the financial market, has been a popular topic in both academia and financial industry since 1970s. In the last decade, reinforcement learning (RL) has garnered significant interest in many domains such as robotics and video games, owing to its outstanding ability on solving complex sequential decision making problems. RL's impact is pervasive, recently demonstrating its ability to conquer many challenging QT tasks. It is a flourishing research direction to explore RL techniques' potential on QT tasks. This paper aims at providing a comprehensive survey of research efforts on RL-based methods for QT tasks. More concretely, we devise a taxonomy of RL-based QT models, along with a comprehensive summary of the state of the art. Finally, we discuss current challenges and propose future research directions in this exciting field.
\end{abstract}



\maketitle

\section{Introduction}
Quantitative trading has been a lasting research area at the intersection of finance and computer science for many decades. In general, QT research can be divided into two directions. In the finance community, designing theories and models to understand and explain the financial market is the main focus. The famous capital asset pricing model (CAPM) \cite{sharpe1964capital}, Markowitz portfolio theory \cite{markowitz1959portfolio} and Fama \& French factor model \cite{fama1993common} are a few representative examples. On the other hand, computer scientists apply data-driven ML techniques to analyze financial data \cite{ding2015deep,patel2015predicting}. Recently, deep learning becomes an appealing approach owing to not only its stellar performance but also to the attractive property of learning meaningful representations from scratch. 

RL is an emerging subfield of ML, which provides a mathematical formulation of learning-based control. With the usage of RL, we can train agents with near-optimal behaviour policy through optimizing task-specific reward functions \cite{sutton2018reinforcement}. In the last decade, we have witnessed many significant artificial intelligence (AI) milestones achieved by RL approaches in domains such as Go \cite{silver2016mastering}, video games \cite{mnih2015human} and robotics \cite{levine2016end}. RL-based methods also have achieved state-of-the-art performance on many QT tasks such as algorithmic trading (AT) \cite{liu2020adaptive}, portfolio management \cite{wang2020commission}, order execution \cite{fang2021universal} and market making \cite{spooner2018market}. It is a promising research direction to address QT tasks with RL techniques.

\subsection{Why Reinforcement Learning for Quantitative Trading?}
In general, the overall objective of QT tasks is to maximize long-term profit under certain risk tolerance. Specifically, algorithmic trading makes profit through consistently buying and selling one given financial asset; Portfolio management tries to maintain a well-balanced portfolio with multiple financial assets; Order execution aims at fulfilling a specific trading order with minimum execution cost; Market making provides liquidity to the market and makes profit from the tiny price spread between buy and sell orders. Traditional QT strategies \cite{jegadeesh1993returns,moskowitz2012time,poterba1988mean} discover trading opportunities based on heuristic rules. Finance expert knowledge is incorporated to capture the underlying pattern of the financial market. However, rule-based methods exhibit poor generalization ability and only perform well in certain market conditions \cite{deng2016deep}. Another paradigm is to trade based on financial prediction. In the literature, there are also attempts using supervised learning methods such as linear models \cite{ariyo2014stock, bhuriya2017stock}, tree-based models \cite{khaidem2016predicting, ke2017lightgbm} and deep neural networks \cite{devadoss2013forecasting, selvin2017stock} for financial prediction. Nevertheless, the high volatility and noisy nature of the financial market make it extremely hard to predict future price accurately \cite{fama2021efficient}. In addition, there is an unignorable gap between prediction signals and profitable trading actions. Thus, the overall performance of prediction-based methods is not satisfying as well. 

To design profitable QT strategies, the advantages of RL methods are four-fold: (i) RL allows training an end-to-end agent, which takes available market information as input state and output trading actions directly. (ii) RL-based methods bypass the extremely difficult task to predict future price and optimize overall profit directly. (iii) Task-specific constraints (e.g., transaction cost and slippage) can be imported into RL objectives easily. (iv) RL methods have the potential to generalize to any market condition.

\subsection{Difference from Existing Surveys}
To the best of our knowledge, this survey will be the first comprehensive survey on RL-based QT applications. Although there are some existing works trying to explore the usage of RL techniques in QT tasks, none of them has provided an in-depth taxonomy of existing works, analyzed current challenges of this research field or proposed future directions in this area. The goal of this survey is to provide a summary of existing RL-based methods for QT applications from both RL algorithm perspective and application domain perspective, to analyze current status of this field, and to point out future research directions.

A number of survey papers on ML in finance have been presented in recent years. For example, \citet{rundo2019machine} proposed a brief survey on ML for QT. \citet{emerson2019trends} focused on the trend and applications. \citet{bahrammirzaee2010comparative} introduced many hybrid methods in financial applications. \citet{gai2018survey} proposed a review of Fintech from both ML and general perspectives. \citet{zhang2004discovering} discussed about data mining approaches in Fintech. \citet{chalup2008kernel} discussed about kernel methods in financial applications. Agent-based computational finance is the focus of \cite{tesfatsion2006handbook}. There are also many surveys on deep learning for finance. \citet{wong1998neural} covered early works. \citet{sezer2020financial} was a recent survey paper with a focus on financial time series forecasting. \citet{ozbayoglu2020deep} made a survey on the development of DL in financial applications. \citet{fischer2018reinforcement} presented a brief review of RL methods in the financial market.   

Considering the increasing popularity and potential of RL-based QT applications, a comprehensive survey will be of high scientific and practical values. More than 100 high quality papers are shortlisted and categorized in this survey. Furthermore, we analyze current situation of this area and point out future research directions.

\subsection{How Do We Collect Papers?}
Google scholar is used as the main search engine to collect relevant papers. In addition, we screened related top conferences such as NeurIPS, ICML, IJCAI, AAAI, KDD, just to name a few, to collect high-quality relevant publications. Major key words we used are reinforcement learning, quantitative finance, algorithmic trading, portfolio management, order execution, market making, and stock.
\subsection{Contribution of This Survey}
This survey aims at thoroughly reviewing existing works on RL-based QT applications. We hope it will provide a panorama, which will help readers quickly get a full picture of research works in this field. In conclusion, the main contribution of this survey is three-fold: (i) We propose a comprehensive survey on RL-based QT applications and categorized existing works from different perspectives. (ii) We analyze the advantages and disadvantages of RL techniques for QT and highlight the pathway of current research. (iii) We discuss current challenges and point out future research directions.

\subsection{Article Organization}
The remainder of this article is organized as follows: Section 2 introduces background of QT. Section 3 provides a brief description of RL. Section 4 discusses the usage of supervised learning (SL) methods in QT. Section 5 makes a comprehensive review of RL methods in QT. Section 6 discusses current challenges and possible open research directions in this area. Section 7 concludes this paper.

\begin{figure}
\begin{center}
\includegraphics[width=0.9\textwidth]{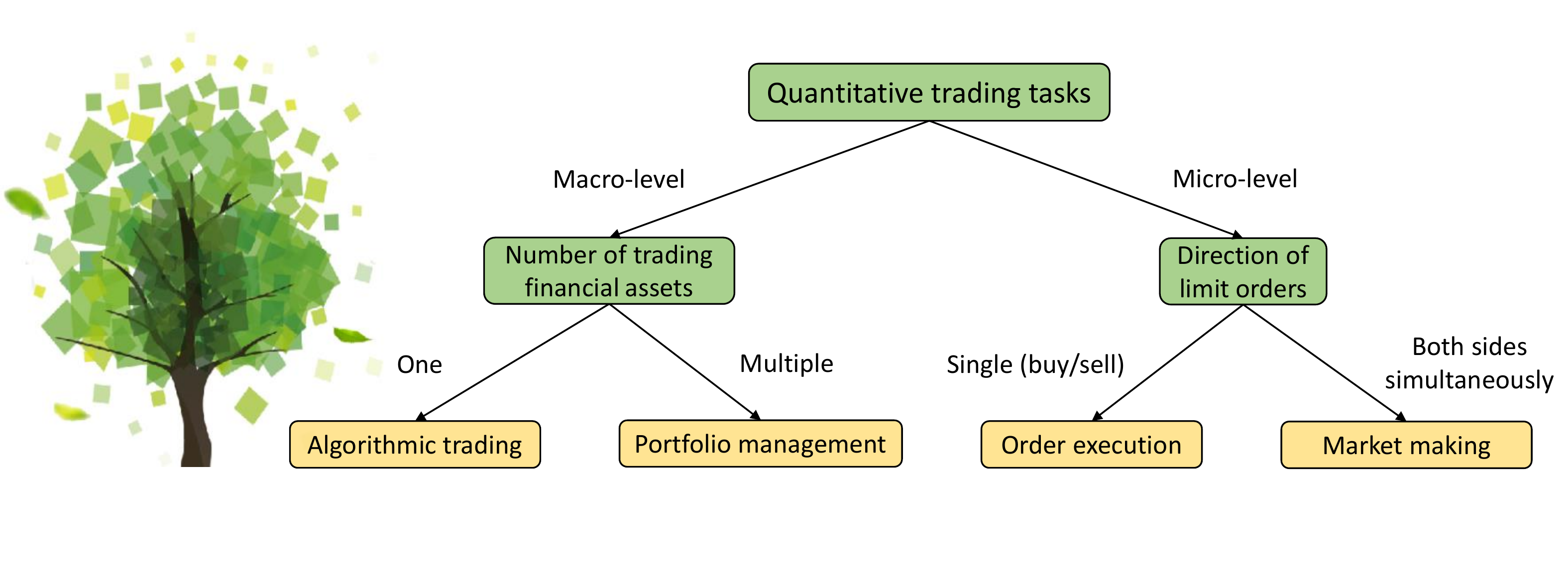}
\caption{Relationships between Quantitative Trading Tasks}
\label{tree}
\end{center}
\end{figure}
\section{Quantitative Trading Background}
Before diving into details of this survey, we introduce background knowledge of QT in this section. Relationships between different QT tasks are illustrated in Fig. \ref{tree}. We propose an overview and then introduce mainstream QT tasks in details respectively. A summary of notations is illustrated in Table. \ref{notation}.

\begin{table}[htbp!]
\newcommand{\tabincell}[2]{\begin{tabular}{@{}#1@{}}#2\end{tabular}}  
\begin{tabular}{|c|l|}
\hline
Notation & Description \\ \hline
\(h\)& length of a holding period \\\hline
\(\mathit{\mathbf{p}}_{i}\) & the time series vector of asset \(i\)'s price \\\hline
\(p_{i,t}\) & the price of asset \(i\) at time \(t\) \\\hline
\(p_{i,t}^{\prime}\) & the price of asset \(i\) after a holding period \(h\) from time \(t\) \\\hline
\(p_{t}\) & the price of a single asset at time \(t\) \\\hline
\(s_t\) & position of an asset at time \(t\) \\\hline
\(u^{i}_{t}\) & trading volume of asset \(i\) at time \(t\) \\\hline
\(\mathbf{n}\) & the time series vector of net value \\\hline
\(n_{t}\) & net value at time \(t\) \\\hline
\(n_{t}^{\prime}\) & net value after a holding period \(h\) from time \(t\) \\\hline
\( w^{i}_{t} \)& portfolio weight of asset \(i\) at time \(t\) \\\hline
\(\mathit{\mathbf{w}_{t}}\)& portfolio vector at time \(t\) \\\hline
\(\mathit{\mathbf{w}_{t}^{\prime}}\)& portfolio vector after a holding period \(h\) from time \(t\) \\\hline
\(v_{t}\) & portfolio value at time \(t\) \\\hline
\(v_{t}^{\prime}\) & portfolio value after a holding period \(h\) from time \(t\) \\\hline
\(f^{i}_{t}\)& transaction fee for asset \(i\) at time \(t\) \\\hline
\( \xi \) & transaction fee rate \\\hline
\(q\) & the quantity of a limit order \\\hline
\(Q\) & total quantity required to be executed \\\hline
\(\mathbf{r}\) & the time series vector of return rate \\\hline
\(r_t\) & return rate at time \(t\) \\ \hline
\end{tabular}
\caption{A Summary of Notations}\label{notation}
\end{table}

\subsection{Overview}
The financial market, an ecosystem involving transactions between businesses and investors, observed a market capitalization exceeding \$80 trillion globally as of the year 2019.\footnote{https://data.worldbank.org/indicator/CM.MKT.LCAP.CD/} For many countries, the financial industry has become a paramount pillar, which spawns the birth of many financial centres. The International Monetary Fund (IMF) categorizes financial centres as follows: international financial centres, such as New York, London and Tokyo; regional financial centres, such as Shanghai, Shenzhen and Sydney; offshore financial centres, such as Hong Kong, Singapore and Dublin. At the core of financial centres, trading exchanges,where trading activities involving trillions of dollars take place everyday, are formed. Trading exchanges can be divided as stock exchanges such as NYSE, Nasdaq and Euronext, derivatives exchanges such as CME and cryptocurrency exchanges such as Coinbase and Huobi. Participants in the financial market can be generally categorized as financial intermediaries (e.g., banks and brokers), issuers (e.g., companies and governments), institutional investors (e.g., investment managers and hedge funds) and individual investors. With the development of electronic trading platform, quantitative trading, which has been demonstrated quite profitable by many leading trading companies (e.g., Renaissance\footnote{https://www.rentec.com/}, Two Sigma\footnote{https://www.twosigma.com/}, Cithadel\footnote{https://www.citadel.com/}, D.E. Shaw\footnote{https://www.deshaw.com/}), is becoming a dominating trading style in the global financial markets. In 2020, quantitative trading accounts for over 70\% and 40\% trading volume in developed market (e.g., US and Europe) and emerging market (e.g., China and India) respectively. \footnote{https://therobusttrader.com/what-percentage-of-trading-is-algorithmic/} We introduce some basic QT concepts as follows:
\begin{itemize}
    \item \(\textbf{Financial \ Asset.}\)
A financial asset refers to a liquid asset, which can be converted into cash immediately during trading time. Classic financial assets include stocks, futures, bonds, foreign exchanges and cryptocurrencies. 

    \item \(\textbf{Holding \ Period.}\)
Holding period \(h\) refers to the time period where traders just hold the financial assets without any buying or selling actions. 

    \item \(\textbf{Asset \ Price.}\)
The price of a financial asset \(i\) is defined as a time series \(\mathit{\mathbf{p_{i}} = \{p_{i,1},p_{i,2},p_{i,3},...,p_{i,t}\}}\), where \(p_{i,t}\) denotes the price of asset \(i\) at time \(t\). \(p_{i,t}^{\prime}\) is the price of asset \(i\) after a holding period \(h\) from time \(t\). \( p_{t}\) is used to denote the price at time \(t\) when there is only one financial asset.

    \item \(\textbf{OHLC.}\) OHLC is the abbreviation of open price, high price, low price and close price. The candle stick, which is consisted of OHLC, is widely used to analyze the financial market.
    
    \item \(\textbf{Volume.}\) Volume is the amount of a financial asset that changes hands. \( u^{i}_{t}\) is the trading volume of asset \(i\) at time \(t\).
    
    \item \(\textbf{Technical \ Indicator.}\) A technical indicator indicates a feature calculated by a formulaic combination of OHLC and volume. Technical indicators are usually designed by finance experts to uncover the underlying pattern of the financial market.

    \item \(\textbf{Return \ Rate.}\)
Return rate is the percentage change of capital, where \(r_t = (p_{t+1}-p_{t})/p_t\) denotes the return rate at time \(t\). The time series of return rate is denoted as \(\mathbf{r}=(r_1,r_2,...,r_t)\).
    
    \item \(\textbf{Transaction \ Fee.}\)
Transaction fee is the expenses incurred during trading financial assets: \(f^{i}_{t} = p_{i,t} \times u^{i}_{t} \times \xi \), where \( \xi \) is the transaction fee rate.

    \item \(\textbf{Liquidity.}\) 
Liquidity refers to the efficiency with which a financial asset can be converted into cash without having an evident impact on its market price. Cash itself is the asset with the most liquidity.

\end{itemize}
 
\begin{table}[htbp!]
\newcommand{\tabincell}[2]{\begin{tabular}{@{}#1@{}}#2\end{tabular}}  
\begin{tabular}{|l|l|l|}
\hline
\textbf{Trading Style} & \textbf{Time Frame} & \textbf{Holding Period} \\ \hline
Position trading & Long term & Months to years \\ \hline
Swing trading & Medium term & Days to weeks \\ \hline
Day trading & Short term & Within a trading day \\ \hline
Scalping trading & Short term & Seconds to minutes \\ \hline
High-frequency trading & Extreme short term & Milliseconds to seconds \\ \hline
\end{tabular}
\caption{A Summary of Algorithmic Trading Styles}\label{style}
\end{table}
\subsection{Algorithmic Trading}
Algorithmic trading (AT) refers to the process that traders consistently buy and sell one given financial asset to make profit. It is widely applied in trading stocks, commodity futures and foreign exchanges. For AT, time is splitted as discrete time steps. At the beginning of a trading period, traders are allocated some cash and set net value as 1. Then, at each time step \(t\), traders have the options to buy, hold or sell some amount of shares for changing positions. Net value and position is used to represent traders' status at each time step. The objective of AT is to maximize the final net value at the end of the trading period. Based on trading styles, algorithmic trading is generally divide 5 categories: position trading, swing trading, day trading, scalp trading and high-frequency trading. Specifically, position trading involves holding the financial asset for a long period of time, which is unconcerned with short-term market fluctuations and only focuses on the overarching market trend. Swing trading is a medium-term style that holds financial assets for several days or weeks. The goal of swing trading is to  spot a trend and then capitalise on dips and peaks that provide entry points. Day trading tries to capture the fleeting intraday pattern in the financial market and all positions will be closed at the end of the day to avoid overnight risk. Scalping trading aims at discovering micro-level trading opportunities and makes profit by holding financial assets for only a few minutes. High-frequency trading is a type of trading style characterized by high speeds, high turnover rates, and high order-to-trade. A summary of different trading styles is illustrated in Table. \ref{style}.       

Traditional AT methods discover trading signals based on technical indicators or mathematical models. Buy and Hold (BAH) strategy, which invests all capital at the beginning and holds until the end of the trading period, is proposed to reflect the average market condition. Momentum strategies, which assumes the trend of financial assets in the past has the tendency to continue in the future, are another well-known AT strategies. Buying-Winner-Selling-Loser \cite{jegadeesh1993returns}, Times Series Momentum \cite{moskowitz2012time} and Cross Sectional Momentum \cite{chan1996momentum} are three classic momentum strategies. In contrast, mean reversion strategies such as Bollinger bands \cite{bollinger2002bollinger} assume the price of financial assets will finally revert to the long-term mean. Although traditional methods somehow capture the underlying patterns of the financial market, these simple rule-based methods exhibit limited generalization ability among different market conditions. We introduce some basic AT concepts as follows: 
\begin{itemize}
    \item \(\textbf{Position.}\)
Position \(s_t\) is the amount of a financial asset owned by traders at time \(t\). It represents a long (short) position when \(s_t\) is positive (negative). 

    \item \(\textbf{Long \ Position.}\)
Long position makes positive profit when the price of the asset increases. For long trading actions, which buy a financial asset \(i\) at time \(t\) first and then sell it at \(t+1\), the profit is \(u^{i}_{t}(p_{i,t+1}-p_{i,t})\), where \(u^{i}_{t}\) is the buying volume of asset \(i\) at time \(t\).

    \item \(\textbf{Short \ Position.}\)
Short position makes positive profit when the price of the asset decreases. For short trading actions, which buys a financial asset at time \(t\) first and then sell it at \(t+1\), the profit is \(u^{i}_{t}(p_{i,t}-p_{i,t+1})\).
   
    \item \(\textbf{Net \ Value.}\)
Net value represents a fund's per share value. It is defined as a time series \( \mathbf{n}=\{n_{1},n_{2},...,n_{t}\} \), where \(n_{t}\) denotes the net value at time \(t\). The initial net value is always set to 1.
\end{itemize}

\subsection{Portfolio Management}
\begin{figure}
\begin{center}
\includegraphics[width=0.6\textwidth]{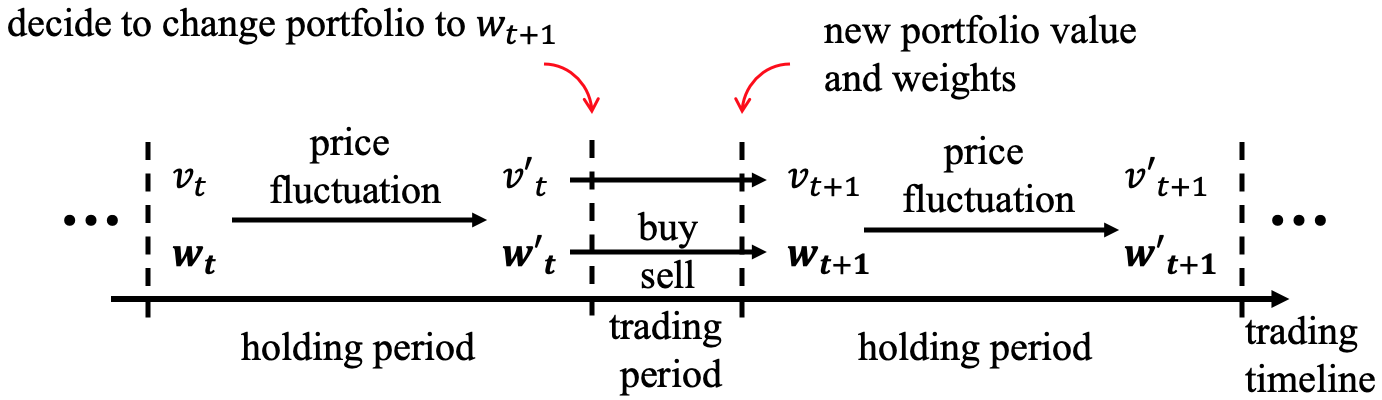}
\caption{Portfolio Management Process}
\label{pm_process}
\end{center}
\end{figure}

Portfolio management (PM) is a fundamental QT task, where investors hold a number of financial assets and reallocate them periodically to maximize long-term profit. In the literature, it is also called portfolio optimization, portfolio selection and portfolio allocation. In the real market, portfolio managers work closely with traders, where portfolio managers assign a percentage weighting to every stock in the portfolio periodically and traders focus on finishing portfolio reallocation at the favorable price to minimize the trading cost. For PM, time is splitted into two types of periods: holding period and trading period as shown in Figure \ref{pm_process}. At the beginning of a holding period, the agent holds a portfolio \(\mathbf{w_{t}}\) consists of pre-selected financial assets with a corresponding portfolio value \(v_t\). With the fluctuation of the market, the assets' prices would change during the holding period. At the end of the holding period, the agent will get a new portfolio value \(v_t^\prime\) and decide a new portfolio weight \(\mathbf{w_{t+1}}\) of the next holding period. During the trading period, the agent buys or sells some shares of assets to achieve the new portfolio weights. The lengths of the holding period and trading period are based on specific settings and can change over time. In some previous works, the trading period is set to 0, which means the change of portfolio weight is achieved immediately for convenience. The objective is to maximize the final portfolio value given a long time horizon.

PM has been a fundamental problem for both finance and ML community for decades. Existing approaches can be grouped into four major categories, which are benchmarks such as Constant Rebalanced Portfolio (CRP) and Uniform Constant Rebalanced Portfolio (UCRP) \cite{cover2011universal}, Follow-the-Winner approaches such as Exponential Gradient (EG) \cite{helmbold1998line} and Winner \cite{gaivoronski2000stochastic}, Follow-the-Loser approaches such as Robust Mean Reversion (RMR) \cite{huang2012robust}, Passive Aggressive Online Learning (PAMR) \cite{li2012pamr} and Anti-Correlation \cite{borodin2004can}, Pattern-Matching-based approaches such as correlation-driven nonparametric learning (CORN) \cite{li2011corn} and \(B^K\) \cite{gyorfi2006nonparametric}, and Meta-Learning algorithms such as Online Newton Step (ONS). The readers can check this survey \cite{li2014online} for more details. We introduce some basic PM concepts as follows: 
\begin{itemize}
    \item \(\textbf{Portfolio.}\)
A portfolio can be represented as:
\[ \mathbf{w_{t}} = {[w^0_{t},w^1_{t},...,w^M_{t}]}^{T} \in R^{M+1} \quad and \quad \sum_{i=0}^{M}w^i_{t} = 1 \]
where M+1 is the number of portfolio's constituents, including one risk-free asset, i.e., cash, and M risky assets. \(w^i_{t}\) represents the ratio of the total portfolio value (money) invested at the beginning of the holding period \(t\) on asset i. Specifically, \(w^0_{t}\) represents the cash in hand.
    \item \(\textbf{Portfolio \ Value.}\)
We define \(v_{t}\) and \(v_{t}^{\prime}\) as portfolio value at the beginning and end of the holding period. So we can get the change of portfolio value during the holding period and the change of portfolio weights:
\[v_{t}^{\prime} = v_{t}\sum_{i=0}^{M}\frac{w^{i}_{t}p_{i,t}^{\prime}}{p_{i,t}}\quad w_{t}^{\prime} = \frac{\frac{w^{i}_{t}p_{i,t}^{\prime}}{p_{i,t}}}{{\textstyle \sum_{i=0}^{M}\frac{w^{i}_{t}p_{i,t}^{\prime}}{p_{i,t}}}}\quad for \quad i \in \left[0,M\right]\]
\end{itemize}

\subsection{Order Execution}
While adjusting new portfolio, investors need to buy (or sell) some amount of shares by executing an order of liquidation (or acquirement). Essentially, the objectives of order execution are two-fold: it does not only require to fulfill the whole order but also target a more economical execution with maximizing profit (or minimizing cost). As mentioned in \cite{cartea2015algorithmic}, the major challenge of order execution lies in the trade-off between avoiding harmful market impact caused by large transactions in a short period and restraining price risk, which means missing good trading windows due to slow execution. Traditional OE solutions are usually designed based on some stringent assumptions of the market and then derive some model-based methods with stochastic control theory. For instance, Time Weighted Average Price (TWAP) evenly splits the whole order and execute at each time step with the assumption that the market price follows the Brownian motion \cite{bertsimas1998optimal}. 
\begin{wrapfigure}{r}{4cm}
    \begin{center}
    \includegraphics[width=0.3\textwidth]{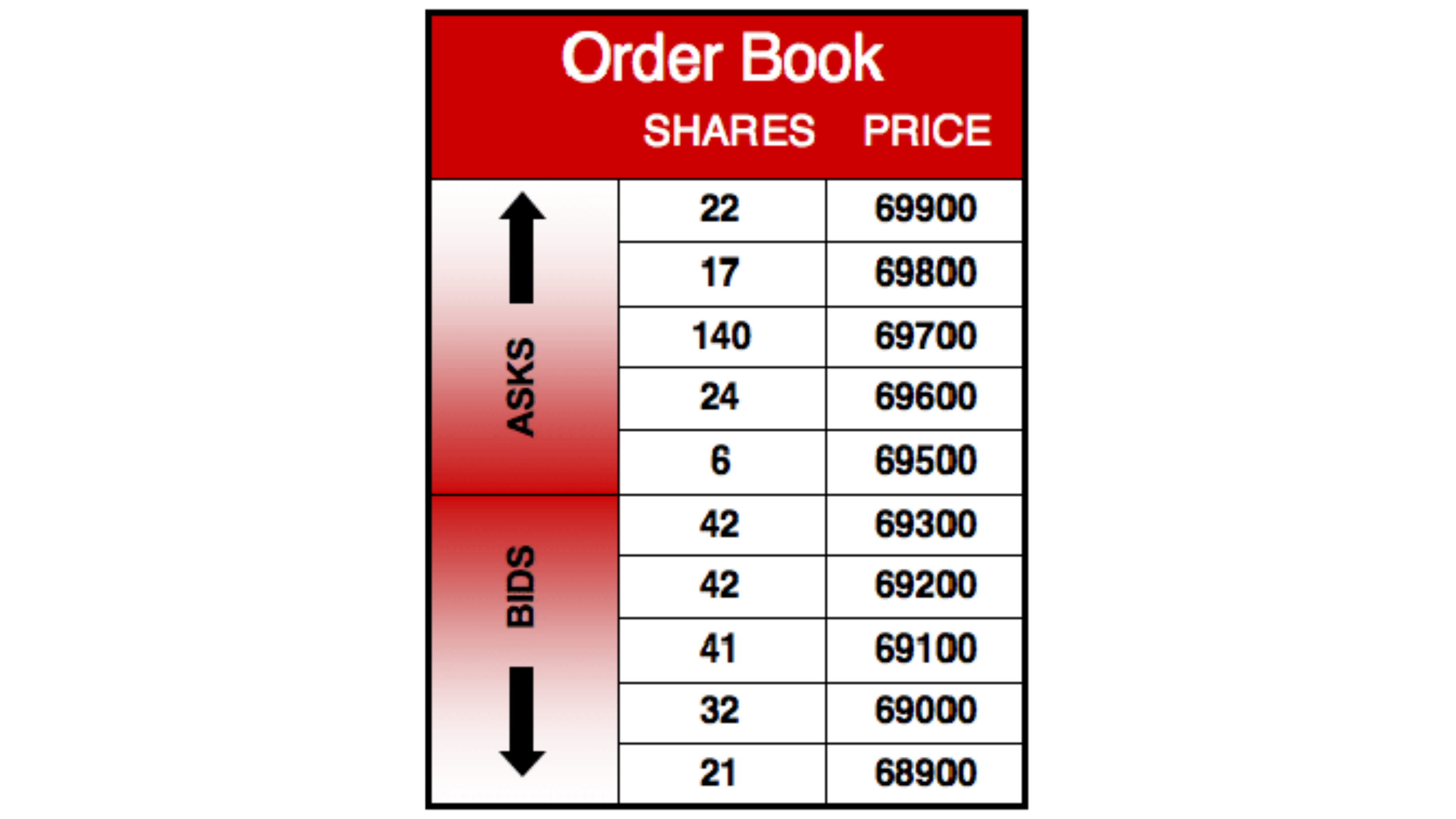}
    \caption{Limit Order Book}
    \label{orderbook}
    \end{center}
\end{wrapfigure}
The Almgren-Chriss model \cite{almgren2001optimal} incorporates temporary and permanent price impact functions also with the Brownian motion assumption. Volume Weighted Average Price (VWAP) distributes orders in proportion to the (empirically estimated) market transaction volume. The goal of VWAP is to track the market average execution price \cite{kakade2004competitive}. However, traditional solutions are not effective in the real market because of the inconsistency between the assumptions and reality.

Formally, OE is to trade fixed amount of shares within a predetermined time horizon (e.g., one hour or one day). At each time step \(t\), traders can propose to trade a quantity of \(q_t \ge 0\) shares at current market price \(p_t\), The matching system will then return the execution results at time \(t+1\). Taking the sell side as an example, assuming a total of Q shares required to be executed during the whole time horizon, the OE task can be formulated as:
\[\text{arg} \max_{{q}_{1},{q}_{2},...,{q}_{T}}  \sum_{t=1}^{T}({q}_{t}\cdot{p}_{t} ),\quad \text{s.t.}\sum_{t=1}^{T}{q}_{t} = Q \]
OE not only completes the liquidation requirement but also the maximize/minimize average execution price for the sell/buy side execution respectively. We introduce basic OE concepts as follows: 

\begin{itemize}
    \item \(\textbf{Market \ Order.}\)
A market order refers submitting an order to buy or sell a financial asset at the current market price, which expresses the desire to trade at the best available price immediately.

    \item \(\textbf{Limit \ Order.}\)
A limit order is an order placed to buy or sell a number of shares at a specified price during a specified time frame. It can be modeled as a tuple \(p_{target}\pm q_{target}\), where \(p_{target}\) represents the submitted target price, \(q_{target}\) represents the submitted target quantity, and \(\pm\) represents trading direction (buy/sell)
    \item \(\textbf{Limit \ Order \ Book.}\)
A limit order book (LOB) is a list containing all the information about the current limit orders in the market. An example of LOB is shown in Figure \ref{orderbook}.
    \item \(\textbf{Average Execution Price}\)
    Average execution price (AEP) is defined as \(\bar{p}= {\textstyle \sum_{t=1}^{T}\frac{{q}_{t}}{Q} }\cdot {p}_{t}\).
    \item \(\textbf{Order \ Matching \ System.}\)
The electronic system that matches buy and sell orders for a financial market is called the order matching system. The matching system is the core of all electronic exchanges, which decides the execution results of orders in the market. The most common matching mechanism is first-in-first-out, which means limit orders at the same price will be executed in the order in which the orders were submitted.
\end{itemize}

\subsection{Market Making}
Market makers are traders who continually quote prices at which they are willing to trade on both buy and sell side for one financial asset. They provide liquidity and make profit from the tiny price spread between buy and sell orders. The main challenge for market making is non-zero inventory. When you submit a limit order on both sides, there is no guarantee that all the orders can be successfully executed. It is risky when non-zero inventory accumulates to a high level because this means market maker will have to close the inventory by current market price, which could lead to a significant loss. In practice, some market makers keep their inventory at low-level to avoid market exposure and only make profits by repeatedly making their quoted spread. On the other hand, some more advanced market makers may choose to hold a non-zero inventory to capture the market trend, while exploiting the quoted spread simultaneously. Traditional finance methods consider market making as a stochastic optimal control problem \cite{cartea2014buy}. Agent-based method \cite{gode1993allocative} and RL \cite{spooner2018market} have also been applied to market making. 

\subsection{Evaluation Metrics}
In this subsection, we discuss common profit metrics, risk metrics and risk-adjusted metrics for evaluation in this field.
\subsubsection{Profit Metrics}
\begin{itemize}
    \item \textbf{Profit rate (PR).} PR is the percent change of net value over time horizon \(h\). The formal definition is: \[PR = (n_{t+h}-n_{t})/n_{t}\]
     \item \textbf{Win rate (WR).} WR evaluates the proportion of trading days with positive profit among all trading days.
\end{itemize}
\subsubsection{Risk Metrics}
\begin{itemize}
    \item \textbf{Volatility (VOL).} VOL is the variance of the return vector \(\mathbf{r}\). It is widely used to measure the uncertainty of return rate and reflects the risk level of strategies. The formal definition is:
    \[VOL = \sigma[\mathbf{r}]\]
    \item \textbf{Maximum drawdown (MDD).} MDD \cite{magdon2004maximum} measures the largest decline from the peak in the whole trading period to show the worst case. The formal definition is:
    \[MDD = \max_{\tau\in(0,t)}  [\max_{t\in(0,\tau)}\frac{n_t-n_\tau}{n_t} ]  \]
    \item \textbf{Downside deviation (DD).} DD refers to the standard deviation of trade returns that are negative.
    \item \textbf{Gain-loss ratio (GLR).} GLR is a downside risk measure. It represents the relative relationship of trades with a positive return and trades with a negative return. The formula is:
    \[ GLR = \frac{\mathbb{E}[\mathbf{r}|\mathbf{r}>0 ] }{\mathbb{E}[-\mathbf{r}|\mathbf{r}<0 ]} \]
\end{itemize}
\subsubsection{Risk-adjusted Metrics}
\begin{itemize}
    \item \textbf{Sharpe ratio (SR).} SR \cite{sharpe1994sharpe} is a risk-adjusted profit measure, which refers to the return per unit of deviation:
    \[SR = \frac{\mathbb{E}[\mathbf{r}]}{\sigma[\mathbf{r}]}\]
    \item \textbf{Sortino ratio (SoR).} SoR is a variant of risk-adjusted profit measure, which applies DD as risk measure:
    \[SoR = \frac{\mathbb{E}[\mathbf{r}]}{DD}\]
    \item \textbf{Calmar ratio (CR).} CR is another variant of risk-adjusted profit measure, which applies MDD as risk measure:
    \[CR = \frac{\mathbb{E}[\mathbf{r}]}{MDD}\]
\end{itemize}

\section{Overview of Reinforcement Learning}
RL is a popular subfield of ML that studies complex decision making problems. \citet{sutton2018reinforcement} distinguish RL problems by three key characteristics: (i) the problem is closed-loop. (ii) the agent figures out what to do through trial-and-error. (iii) actions have an impact on both short term and long term results. The decision maker is called agent and the environment is everything else except the agent. At each time step, the agent obtains some observations of the environment, which is called state. Later on, the agent takes an action based on the current state. The environment will then return a reward and a new state to the agent. Formally, an RL problem is typically formulated as a Markov decision process (MDP) in the form of a tuple \( \mathcal{M} = (S, A, R, P, \gamma ) \),  where \(S\) is a set of states \( s\in S \), \( A \) is a set of actions \( a\in A \), \(R\) is the reward function, \(P\) is the transition probability, and \(\gamma\) is the discount factor. The goal of an RL agent is to find a policy \(\pi(a\mid s)\) that takes action \(a\in A\) in state \(s \in S\) in order to maximize the expected discounted cumulative reward:
\[ \max \ \mathbb{E}[R(\tau )], \ where \ R(\tau) = \sum_{t=0}^{\tau}{\gamma}^{t}r({a}_{t},{s}_{t}) \ and \ 0\le \gamma \le 1  \]

\citet{sutton2018reinforcement} summarise RL's main components as: (i) policy, which refers to the probability of taking action \(a\) when the agent is in state \(s\). From policy perspective, RL algorithms are categorized into on-policy and off-policy methods. The goal of on-policy RL methods is to evaluate or improve the policy, which they are now using to make decisions. As for off-policy RL methods, they aim at improving or evaluating the policy that is different from the one used to generate data. (ii) reward: after taking selected actions, the environment sends back a numerical signal reward to inform the agent how good or bad are the actions selected. (iii) value function, which means the expected return if the agent starts in that state \(s\) or state-action pair \((s,a)\), and then acts according to a particular policy \(\pi\) consistently. Value function tells how good or bad your current position is in the long run. (iv) model, which is an inference about the behaviour of the environment in different states.

Plenty of algorithms have been proposed to solve RL problems. Tabular methods and approximation methods are two mainstream directions. For tabular algorithms, a table is used to represent the value function for every action and state pair. The exact optimal policy can be found through checking the table. Due to the curse of dimensionality, tabular methods only work well when the action and state space is small. Dynamic programming (DP), Monto Carlo (MC) and temporal difference (TD) are a few widely studied tabular methods. Under perfect model of environment assumption, DP uses a value function to search for good policies. Policy iteration and value iteration are two classic DP algorithms. MC methods try to learn good policies through sample sequences of states, actions, and reward from the environment. For MC methods, the assumption of perfect environment understanding is not required. TD methods are a combination of DP and MC methods. While they do not need a model from the environment, they can bootstrap, which is the ability to update estimates based on other estimates. From this family, Q-learning \cite{watkins1992q} and SARSA \cite{rummery1994line} are popular algorithms, which belong to off-policy and on-policy methods respectively. 

On the other hand, approximation methods try to find a great approximate function with limited computation. Learning to generalize from previous experiences (already seen states) to unseen states is a reasonable direction. Policy gradient methods are popular approximate solutions. REINFORCE \cite{williams1992simple} and actor-critic \cite{konda2000actor} are two important examples. With the popularity of deep learning, RL researchers use neural networks as function approximator. DRL is the combination of DL and RL, which lead to great success in many domains \cite{mnih2015human,vinyals2019grandmaster}. Popular DRL algorithms for QT community include deep Q-network (DQN) \cite{mnih2015human}, deterministic policy gradient (DPG) \cite{silver2014deterministic}, deep deterministic policy gradient (DDPG) \cite{lillicrap2015continuous}, proximal policy optimization (PPO) \cite{schulman2017proximal}. More details for RL can be found in \cite{sutton2018reinforcement}. Recurrent reinforcement learning (RRL) is another widely used RL approach for QT. "Recurrent" means the previous output is fed into the model as part of the input here. RRL achieves more stable performance when exposed to noisy data such as financial data.

\section{Supervised Learning for Quantitative Trading}
Supervised learning techniques have been widely used in the pipeline of QT research. In this section, we propose a brief review of research efforts on supervised learning for QT. We introduce existing works from three perspectives: feature engineering, financial forecasting and enhancing traditional methods with ML.

\subsection{Feature Engineering}
Discovering a series of high-quality features is the foundation of ML algorithms' success. In QT, features, which have the ability to explain and predict future price are also called indicators or alpha factors. Traditionally, alpha factors are designed and tested by finance experts based on domain knowledge. However, this way of mining alpha is very costly and not realistic for individual investors. There are many attempts to automatically discover alpha factors. Alpha101 \cite{kakushadze2016101} introduced a set of 101 public alpha factors. Autoalpha \cite{zhang2020autoalpha} combined genetic algorithm and principle component analysis (PCA) to search for alpha factors with low correlation. ADNN \cite{fang2019alpha} proposed an alpha discovery neural network framework for mining alpha factors. In general, it is harmful to feed all available features into ML models directly. Feature selection approaches are applied to reduce irrelevant and redundant features in QT applications \cite{zhang2020doubleensemble,tsai2010combining,lee2009using}. Another paradigm is to use dimension reduction techniques such as PCA \cite{wold1987principal} and
latent Dirichlet allocation (LDA) \cite{tharwat2017linear} to extract meaningful features. 

\subsection{Financial Forecasting}
The usage of supervised learning methods in financial forecasting is pervasive. Researchers formulate return prediction as a regression task and price trend prediction as a classification task. Linear models such as linear regression \cite{bhuriya2017stock}, LASSO \cite{panagiotidis2018determinants}, elastic net \cite{wu2014nonnegative} are used for financial prediction. Non-linear models including random forest \cite{khaidem2016predicting}, decision tree \cite{basak2019predicting}, support vector machine (SVM) \cite{huang2005forecasting} and LightGBM \cite{sun2020novel} outperform linear models owing to their ability to learn non-linear relationships between features. In recent years, deep learning models including multi-layer perceptron (MLP) \cite{devadoss2013forecasting}, recurrent neural network (RNN) \cite{selvin2017stock}, long short state memory (LSTM) \cite{selvin2017stock}, convolutional neural network (CNN) \cite{hoseinzade2019cnnpred} are prevailing owing to its outstanding ability to learn hidden relationship between features. 

Besides different ML models, there is also a trend to utilize alternative data for improving prediction performance. For instance, economic news \cite{hu2018listening}, frequency of prices \cite{zhang2017stock}, social media \cite{xu2018stock}, financial events \cite{ding2016knowledge}, investment behaviors \cite{chen2019investment} and weather information \cite{zhou2020domain} have been used as extra information to learn intrinsic pattern of financial assets. Graph neural networks have been introduced to model the relationship between stocks \cite{chen2018incorporating,li2020modeling,xu2021rest,sawhney2021exploring}. Hybrid methods are also proposed to further improve prediction performance \cite{huang2012hybrid,liumulti}. 

\subsection{Enhancing Traditional Methods with ML}
Another research direction is to enhance traditional rule-based methods with ML techniques. \citet{lim2019enhancing} enhanced time-series momentum with deep learning. \citet{takeuchi2013applying} applied NN to enhance cross section momentum. \citet{chauhan2020uncertainty} took account uncertainty and look-ahead based on factor models. Alphastock \cite{wang2019alphastock} proposed a deep reinforcement attention network to improve the classic buying-winners-and-selling-losers strategy \cite{jegadeesh1993returns}. \citeauthor{gu2020empirical} explore ML techniques' ability on asset pricing. In \cite{gu2021autoencoder}, an autoencoder architecture was proposed for asset pricing. Compared to pure ML methods, these methods keep the original financial insight and have better explainability.  

Even though supervised ML methods achieve great success in financial forecasting with the combination of feature engineering techniques, there is still an unignorable gap between accurate prediction and profitable trading actions. RL methods can tackle this obstacle through learning an end-to-end agent, which maps market information into trading actions directly. In the next section, we will discuss notable RL methods for QT tasks and why they are superior to traditional methods.

\begin{table}[!htbp]
\newcommand{\tabincell}[2]{\begin{tabular}{@{}#1@{}}#2\end{tabular}}  
\begin{tabular}{|c|c|l|}
\hline
Category & RL algorithm & Publication \\ \hline
 & Q-learning & \cite{gao2000algorithm,lee2002multi,jangmin2006adaptive,bertoluzzo2012testing,neuneier1996optimal,neuneier1998enhancing,nevmyvaka2006reinforcement,hendricks2014reinforcement,spooner2018market,lim2018reinforcement,zhongdata}  \\
Value-Based & SARSA & \cite{de2020tabular,chan2001electronic,spooner2018market,spooner2020robust} \\ 
 & DQN & \cite{jeong2019improving,wang2020commission,ning2018double,daberius2019deep,zhang2020deep,yuan2020using} \\ \hline
& RRL & \cite{moody1997optimization,moody1998performance,moody2001learning,dempster2006automated,deng2016deep,si2017multi,ding2018investor,shi2019multi,xu2020relation} \\
Policy-Based & REINFORCE & \cite{wang2020commission} \\
 & PG & \cite{zhang2020deep,benhamou2020bridging,zhang2020cost,wang2021deeptrader,liang2018adversarial} \\
 & TRPO & \cite{vittori2020option,bisi2019risk} \\ \hline
 & DPG & \cite{liu2020adaptive,jiang2017deep,jiang2017cryptocurrency,ye2020reinforcement} \\
 & PPO & \cite{daberius2019deep,linend,fang2021universal,yuan2020using,liang2018adversarial,yang2020deep} \\
Actor-Critic & DDPG & \cite{xiong2018practical,sawhney2021quantitative,liang2018adversarial,yang2020deep} \\
 & SAC &  \cite{yuan2020using}\\
 & A2C & \cite{zhang2020deep,yang2020deep} \\ \hline
Others & Model-based RL & \cite{yu2019model,gueant2019deep} \\
 & Multi-Agent RL & \cite{lee2002multi,lee2020maps,huang2021modularized} \\ \hline
\end{tabular}
\caption{Publications Based on Different Reinforcement Learning Algorithms}\label{rl_classification}
\end{table}
\section{Reinforcement learning for quantitative trading}
In this section, we present a comprehensive review of notable RL-based methods for QT. We go through existing works across four mainstream QT tasks with a summary table at the end of each subsection. 

\subsection{Categories of RL-based QT models}
\begin{figure}[!bht]
  \centering
     \subfloat[{Financial Markets}\label{fig:market}]{%
      \includegraphics[width=0.3\textwidth]{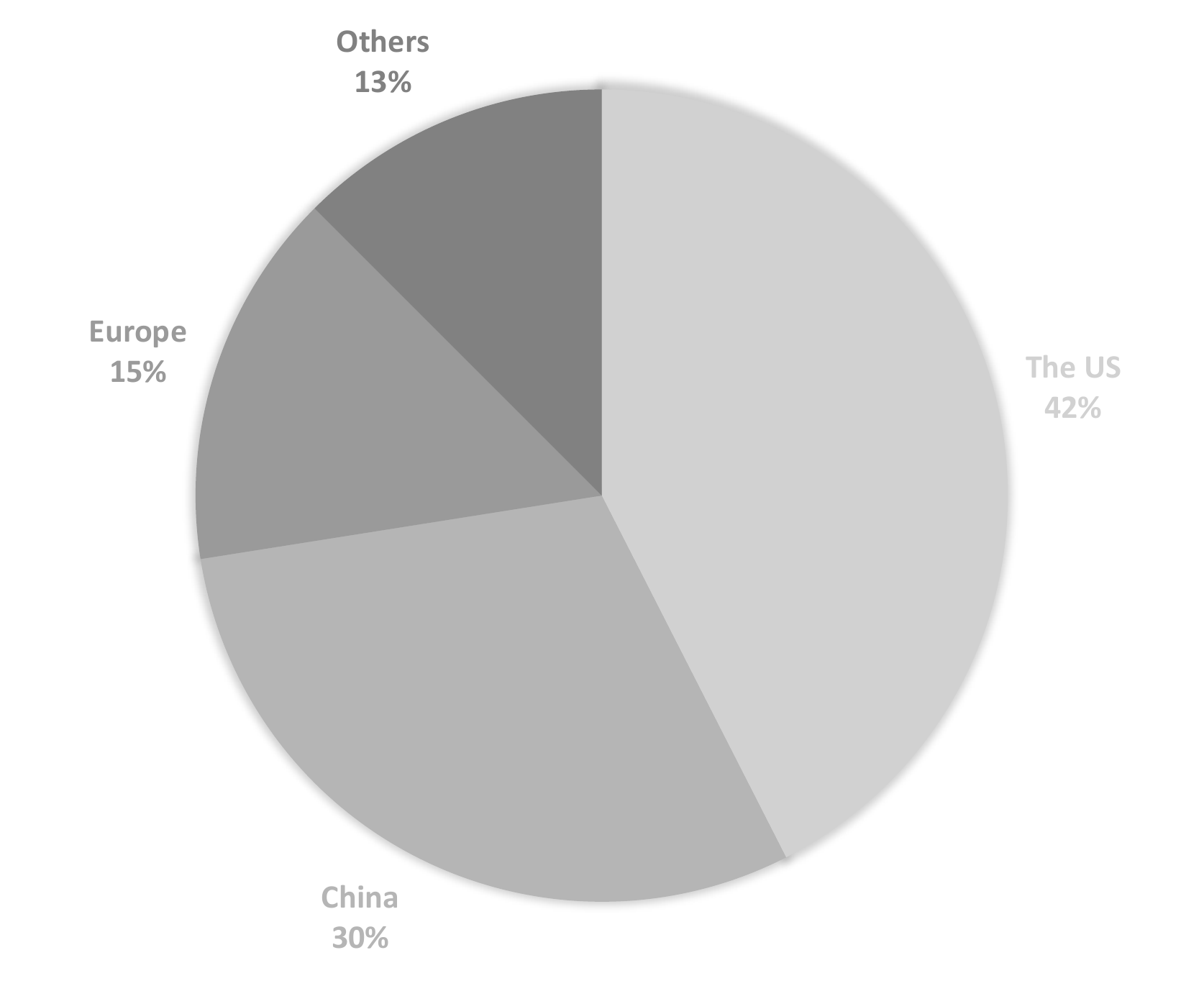}
     }
     \hfill
     \subfloat[{Financial Assets}\label{fig:asset}]{%
      \includegraphics[width=0.3\textwidth]{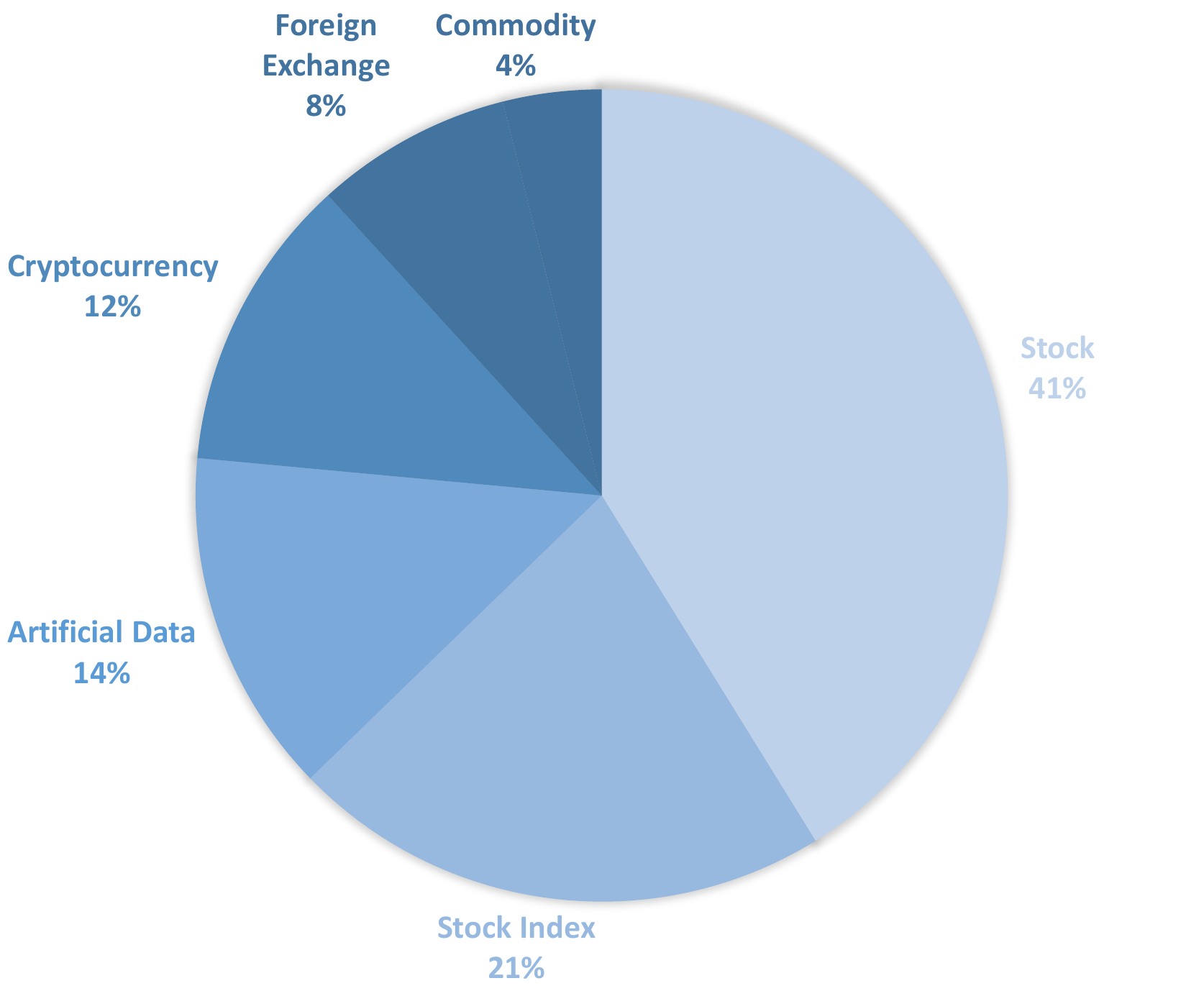}
     }
     \hfill
     \subfloat[{Data Frequencies}\label{fig:frequency}]{%
      \includegraphics[width=0.3\textwidth]{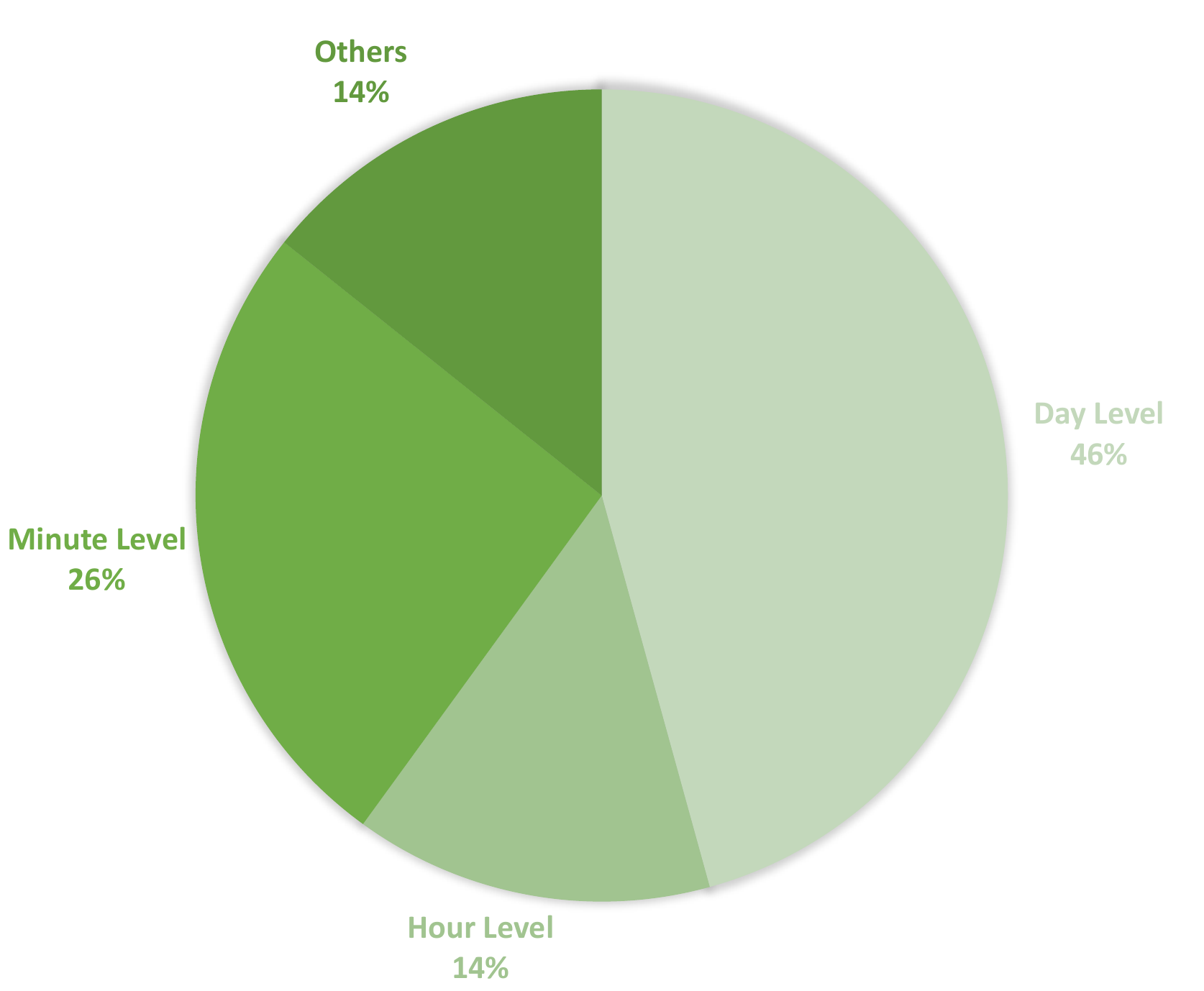}
     }
     \hfill
     \caption{Categorization of  Existing Works}
     \label{fig:trading_behavior}
\end{figure}

In order to provide a bird-eye's view of this field, existing works are classified from different perspectives. Table \ref{rl_classification} summarizes existing works from the RL algorithm perspective. Q-learning and recurrent RL are the most popular RL algorithms for QT. Recent trend indicates DRL methods such as DQN, DDPG and PPO outperform traditional RL methods. In addition, we use three pie charts to provide taxonomies of existing works based on financial markets, financial assets and data frequencies. The percentage numbers shown in the pie charts are calculated by dividing the number of papers belonging to each type with the total number of papers. We classify existing works based on financial markets (illustrated in Figure. \ref{fig:market}). The US market is the most studied market in the literature. Chinese market is getting popular in recent years. The study of European market is mainly in the early era. We classify existing works based on financial assets (illustrated in Figure. \ref{fig:asset}). Stock data is used for more than 40\% of publications. Stock index is the second popular additional option. There are also some works focusing on cryptocurrency in recent years. We classify existing works based on data frequencies (illustrated in Figure. \ref{fig:frequency}). About half of papers use day-level data since it is easy to access. For order execution and market making, fine-grained data (e.g., second-level and minute-level) are often used to simulate the micro-level financial market. 

\subsection{RL in Algorithmic Trading}
Algorithmic trading refers to trade one particular financial asset with signals generated automatically by computer programs. It has been widely applied in trading all kinds of financial assets. In this subsection, we will present a review of most RL-based algorithmic trading papers dating back to 1990s.

\textbf{Policy-based methods.} To tackle the limitations of supervised learning methods, \citet{moody1997optimization} made the first attempt to apply RL in algorithmic trading. In this paper, an agent is trained with recurrent RL (RRL) to optimize the overall profit directly. A novel evaluation metrics called Differential Sharpe Ratio is designed as the optimization objective to improve the performance. Empirical study on artificial price data shows that it outperforms previous forecasting-based methods. Based on the same algorithm, further experiments are conducted using monthly S\&P stock index data  \cite{moody1998performance} and US Dollar/British Pound exchange data \cite{moody2001learning}. As an extension of RRL \cite{moody1997optimization}, \citet{dempster2006automated} proposed an adaptive RRL framework with three layers. Layer one adds 14 technical indicators as extra market information. Layer two evaluates trading actions from layer one with consideration of risk factors. The goal of layer three is to search for optimal values of hyperparameters in layer two. With the three-layer architecture, it outperforms baselines on Eur/ US Dollar exchange data. \citet{vittori2020option} proposed a risk-averse algorithm called Trust Region Volatility Optimization (TRVO) for option hedging. TRVO trains a sheaf of agents characterized by different risk aversion methods and is able to span an efficient frontier on the volatility-p\&l space. Simulation results demonstrate that TRVO outperforms the classic Black \& Scholes delta hedge \cite{black1973pricing}.  

With the development of deep learning, a few DRL methods are proposed for algorithmic trading. FDDR \cite{deng2016deep} enhanced the classic RRL method \cite{moody1998performance} with deep neural networks. An RNN layer is used to learn meaningful recurrent representations of the market. In addition, a fuzzy extension is proposed to further reduce the uncertainty. FDDR achieves great performance on both stock index and commodity futures. To balance between profit and risk, a multi-objective RL method with LSTM layers \cite{si2017multi} is proposed. Through optimizing profit and Sharpe Ratio simultaneously, the agent achieves better performance on 3 Chinese stock index futures. 

\textbf{Value-based methods.} QSR \cite{gao2000algorithm} uses Q-learning to optimize absolute profit and relative risk-adjusted profit respectively. A combination of two networks is employed to improve performance on US Dollar/German Deutschmark exchange data. \citet{lee2002multi} proposed a multi-agent Q-learning framework for stock trading. Four cooperative agents are designed to generate trading signals and order prices for both buy and sell side. Through sharing training episodes and learned policies with each other, this method achieves better performance in terms of both profit and risk management on the Korea stock market compared to supervised learning baselines. In \cite{jangmin2006adaptive}, the authors firstly design some local traders based on dynamic programming and heuristic rules. Later on, they apply Q-learning to learn a meta policy of these local traders on Korea stock markets. \citet{de2020tabular} implemented a SARSA-based RL method and tested it on 10 stocks in the Brazil market.

DQN is used to enhance trading systems by considering trading frequencies, market confusion and transfer learning \cite{jeong2019improving}. The trading frequency is determined in 3 ways: (1) a heuristic function related to Q-value, (2) an action-dependent NN regressor, and (3) an action-independent NN regressor. Another heuristic function is applied to add a filter as the agent's certainty on market condition. Moreover, the authors train the agent on selected component stocks and apply the pre-train weights as the starting point for different stock indexes. Experiments on 4 different stock indexes demonstrate the effectiveness of the proposed framework. 

\textbf{Other methods.} iRDPG \cite{liu2020adaptive} is an adaptive DPG-based framework. Due to the noisy nature of financial data, the authors formulate algorithmic trading as a Partially Observable Markov Decision Process (POMDP). GRU layers are introduced in iRDPG to learn recurrent market embedding. In addition, the authors apply behavior cloning with expert trading actions to guide iRDPG and achieve great performance on two Chinese stock index futures. There are also some works focusing on evaluating the performance of different RL algorithms on their own data. \citet{zhang2020deep} evaluated DQN, PG and A2C on the 50 most liquid futures contracts. \citet{yuan2020using} tested PPO, DQN and SAC on three selected stocks. Based on these two works, DQN achieves the best overall performance among different financial assets. 

\textbf{Summary.} Although existing works demonstrate the potential of RL for quantitative trading, there is seemingly no consensus on a general ranking of different RL algorithms (notably, we acknowledge that \textit{no free lunch theorem} exists). The summary of algorithmic trading publications is in Table \ref{at_2}. In addition, most existing RL-based works only focus on general AT, which tries to make profit through trading one asset. In finance, extensive trading strategies have been designed based on trading frequency (e.g., high-frequency trading) and asset types (e.g., stock and cryptocurrency).

\begin{table}[htbp!]
\newcommand{\tabincell}[2]{\begin{tabular}{@{}#1@{}}#2\end{tabular}}  
\begin{tabular}{cccccc}
\hline
Reference & RL method & Data Source & Asset Type & Market  & Data frequency \\ \hline
\cite{moody1997optimization} & RRL & - & Artificial & -  & - \\ \hline
\cite{moody1998performance} & RRL & - & Stock Index & USA & 1 Month \\ \hline
\cite{gao2000algorithm} & Q-learning & Hand-crafted & FX & - & 1 Day \\ \hline
\cite{moody2001learning} & RRL & Lagged Return & FX & - & 30 Min \\ \hline
\cite{lee2002multi} & Multi-agent RL & Hand-crafted & Stock Index & Korea & - \\ \hline
\cite{jangmin2006adaptive} & Q-learning & Hand-crafted & Stock Index & Korea & - \\ \hline
\cite{dempster2006automated} & RRL & \tabincell{c}{Lagged Return \\ Technical Indicator} & FX & - & 1 Min \\ \hline
\cite{bertoluzzo2012testing} & Q-learning & Lagged Return & \tabincell{c}{Artificial \\ Stock} & \tabincell{c}{- \\ Italy} & \tabincell{c}{- \\ 1 Day} \\ \hline
\cite{deng2016deep} & RRL & Price & \tabincell{c}{Stock Index \\ Commodity} & China & 1 Min \\ \hline
\cite{si2017multi} & RRL & Lagged Return & Stock Index & China & 1 Min \\ \hline
\cite{jeong2019improving} & DQN & Lagged Return & Stock Index & \tabincell{c}{USA, Hong Kong \\ Europe, Korea} & 1 Day \\ \hline
\cite{de2020tabular} & SARSA & OHLC, Technical Indicator & Stock & Brazil & 15 Min \\ \hline
\cite{liu2020adaptive} & DPG & OHLC, Technical Indicator & Stock Index & China & 1 Min \\ \hline
\cite{vittori2020option} & TRPO & Hand-crafted & Artificial & - & - \\ \hline
\cite{zhang2020deep} & DQN, PG, A2C & \tabincell{c}{Price, Lagged Return \\ Technical Indicators} & \tabincell{c}{Stock Index \\ Commodity} & - & - \\ \hline
\cite{yuan2020using} & PPO, DQN, SAC & OHLC & Stock & China & 1 Day \\ \hline

\end{tabular}
\caption{Summary of RL for Algorithmic Trading}\label{at_2}
\end{table}

\subsection{RL in Portfolio Management}
Portfolio management, which studies the art of balancing between a collection of different financial assets, has become a popular topic for RL researchers. In this subsection, we will survey on most notable existing works on RL-based portfolio management.

\textbf{Policy-based methods.} Since a portfolio is essentially a weight distribution among different financial assets, policy-based methods are the most widely applied RL methods for PM. \citet{almahdi2017adaptive} proposed an RRL-based algorithm for portfolio management. Maximum drawdown is applied as the objective function to measure downside risk. In addition, an adaptive version is designed with a transaction cost and market condition stop-loss retraining mechanism. In order to extract information from historical trading records, Investor-Imitator \cite{ding2018investor} formalizes the trading knowledge by imitating the behavior of an investor with a set of logic descriptors. Moreover, to instantiate specific logic descriptors, the authors introduce a Rank-Invest model that can keep the diversity of different logic descriptors through optimizing a variety of evaluation metrics with RRL. Investor-Imitator attempts to imitate 3 types of investors (oracle investor, collaborator investor, public investor) by designing investor-specific reward function for each type. In the experiments on the Chinese stock market, Investor-imitator successfully extracts interpretable knowledge of portfolio management that can help human traders better understand the financial market. Alphastock \cite{wang2019alphastock} is another policy-based RL method for portfolio management. LSTM with history state attention model is used to learn better stock representation. A cross-asset attention network (CAAN) incorporating price rising rank prior is added to further describe the interrelationships among stocks. Later on, the output of CAAN (winning score of each stock) is feed into a heuristic portfolio generator to construct the final portfolio. Policy gradient is used to optimize the Sharpe Ratio. Experiments on both U.S. and Chinese stock market show that Alphastock achieves robust performance over different market states. \({EI}^{3}\) \cite{shi2019multi} is another RRL-based method, which tries to build profitable cryptocurrency portfolios by extracting multi-scale patterns in the financial market. Inspired by the success of Inception networks \cite{szegedy2015going}, the authors design a multi-scale temporal feature aggregation convolution framework with two CNN branches to extract short-term and mid-term market embedding and a max pooling branch to extract the highest price information. To bridge the gap between the traditional Markowitz portfolio and RL-based methods, \citet{benhamou2020bridging} applied PG with a delayed reward function and showed better performance than the classic Markowitz efficient frontier.

\citet{zhang2020cost} proposed a cost-sensitive PM framework based on direct policy gradient. To learn more robust market representation, a novel two-stream portfolio policy network is designed to extract both price series pattern and the relationship between different financial assets. In addition, the authors design a new cost-sensitive reward function to take the trading cost constrain into consideration with theoretically near-optimal guarantee. Finally, the effectiveness of the cost-sensitive framework is demonstrated on real-world cryptocurrency datasets. \citet{xu2020relation} proposed a novel relation-aware transformer (RAT) under the classic RRL paradigm. RAT is structurally innovated to capture both sequential patterns and the inner corrections between financial assets. Specifically, RAT follows an encoder-decoder structure, where the encoder is for sequential feature extraction and the decoder is for decision making. Experiments on 2 cryptocurrency and 1 stock datasets not only show RAT's superior performance over existing baselines but also demonstrate that RAT can effectively learn better representation and benefit from leverage operation. \citet{bisi2019risk} derived a PG theorem with a novel objective function, which exploited the mean-volatility relationship. The new objective could be used in actor-only algorithms such as TRPO with monotonic improvement guarantees. \citet{wang2021deeptrader} proposed DeepTrader, a PG-based DRL method, to tackle the risk-return balancing problem in PM. The model simultaneously uses negative maximum drawdown and price rising rate as reward functions to balance between profit and risk. The authors propose an asset scoring unit with graph convolution layer to capture temporal and spatial interrelations among stocks. Moreover, a market scoring unit is designed to evaluation the market condition. DeepTrader achieves great performance across three different markets.

\textbf{Actor-critic methods.} \citet{jiang2017deep} proposed a DPG-based RL framework for portfolio management. The framework consists of 3 novel components: 1) the Ensemble of Identical Independent Evaluators (EIIE) topology; 2) a Portfolio Vector Memory (PVM); 3) an Online Stochastic Batch Learning (OSBL) scheme. Specifically, the idea of EIIE is that the embedding concatenation of output from different NN layers can learn better market representation effectively. In order to take transaction costs into consideration, PVM uses the output portfolio at the last time step as part of the input of current time step. The OSBL training scheme makes sure that all data points in the same batch are trained in the original time order. To demonstrate the effectiveness of proposed components, extensive experiments using different NN architectures are conducted on cryptocurrency data. Later on, more comprehensive experiments are conducted in an extended version \cite{jiang2017cryptocurrency}. To model the data heterogeneity and environment uncertainty in PM, \citet{ye2020reinforcement} proposed a State-Augmented RL (SARL) framework based on DPG. SARL learns the price movement prediction with financial news as additional states. Extensive experiments on both cryptocurrency and U.S. stock market validation that SARL outperforms previous approaches in terms of return rate and risk-adjusted criteria. Another popular actor-critic RL method for portfolio management is DDPG. \citet{xiong2018practical} constructed a highly profitable portfolio with DDPG on the Chinese stock market. PROFIT \cite{sawhney2021quantitative} is another DDPG-based approach that makes time-aware decisions on PM with text data. The authors make use of a custom policy network that hierarchically and attentively learns time-aware representations of news and tweets for PM, which is generalizable among various actor-critic RL methods. PROFIT shows promising performance on both China and U.S. stock markets. 

\textbf{Other methods.} \citet{neuneier1996optimal} made an attempt to formalize portfolio management as an MDP and trained an RL agent with Q-learning. Experiments on German stock market demonstrate its superior performance over heuristic benchmark policy. Later on, a shared value-function for different assets and model-free policy-iteration are applied to further improve the performance of Q-learning in \cite{neuneier1998enhancing}. There are a few model-based RL methods that attempt to learn some models of the financial market for portfolio management. \cite{yu2019model} proposed the first model-based RL framework for portfolio management, which supports both off-policy and on-policy settings. The authors design an Infused Prediction Module (IPM) to predict future price, a Data Augmentation Module (DAM) with recurrent adversarial networks to mitigate the data deficiency issue, and a Behavior Cloning Module (BCM) to reduce the portfolio volatility. \citet{wang2020commission} focused on a more realistic PM setting where portfolio managers assign a new portfolio periodically for a long-term profit, while traders care about the best execution at the favorable price to minimize the trading cost. Motivated by this hierarchy scenario, a hierarchical RL system (HRPM) is proposed. The high level policy was trained by REINFORCE with an entropy bonus term to encourage portfolio diversification. The low level framework utilizes the branching dueling Q-Network to train agents with 2 dimensions (price and quantity) action space. Extensive experiments are conducted on both US and China stock market to demonstrate the effectiveness of HRPM.

Portfolio management is also formulated as a multi-agent RL problem. MAPS \cite{lee2020maps} is a cooperative multi-agent RL system in which each agent is an independent "investor" creating its own portfolio. The authors design a novel loss function to guide each agent to act as diversely as possible while maximizing its long-term profit. MAPS outperforms most of baselines with 12 years of U.S. stock market data. In addition, the authors find that adding more agents to MAPS can lead to a more diversified portfolio with higher Sharpe Ratio. MSPM \cite{huang2021modularized} is a multi-agent RL framework with a modularized and scalable architecture for PM. MSPM consists of the Evolving Agent Module (EAM) to learn market embedding with heterogeneous input and the Strategic Agent Module (SAM) to produce profitable portfolios based on the output of EAM. 

Some works compare the profitability of portfolios constructed by different RL algorithms on their own data. \citet{liang2018adversarial} compared the performance of DDPG, PPO and PG on Chinese stock market. \citet{yang2020deep} firstly tested the performance of PPO, A2C and DDPG on the U.S. stock market. Later on, the authors find that the ensemble strategy of these three algorithms can integrate the best features and shows more robust performance adjusting to different market situations.

\textbf{Summary.} Since a portfolio is a vector of weights for different financial assets, which naturally corresponds to a policy, policy-based methods are the most widely-used RL methods for PM. There are also many successful examples based on actor-critic algorithms. The summary of portfolio management publications is in Table \ref{pm_2}. We point out two issues of existing methods: (1) Most of them ignore the interrelationship between different financial assets, which is valuable for human portfolio managers. (2) Existing works construct portfolios from a relative small pool of stocks (e.g., 20 in total). However, the real market contains thousands of stocks and common RL methods are vulnerable when the action space is very large \cite{dulac2015deep}.

\begin{table}[htbp!]
\newcommand{\tabincell}[2]{\begin{tabular}{@{}#1@{}}#2\end{tabular}}  
\begin{tabular}{cccccc}
\hline
Reference & RL method & Data Source & Asset Type & Market  & Data frequency \\ \hline
\cite{neuneier1996optimal} & Q-learning & Technical Indicator & Stock Index & Germany & 1 Day \\
\hline
\cite{neuneier1998enhancing} & Q-learning & Hand-crafted & Stock Index & Germany & 1 Day \\ \hline
\cite{jiang2017deep} & DPG & Price, Hand-crafted & Cryptocurrency & - & 30 Min \\ \hline
\cite{almahdi2017adaptive} & RRL & - & Stock Index & - & - \\ \hline
\cite{jiang2017cryptocurrency} & DPG & Lagged Portfolio & Cryptocurrency & - & 30 Min \\ \hline
\cite{liang2018adversarial} & DDPG, PPO, PG & \tabincell{c}{OHLC \\ Technical Indicator} & Stock & China & 1 Day \\ \hline
\cite{ding2018investor} & RRL & Hand-crafted & Stock & China & 1 Day \\ \hline
\cite{xiong2018practical} & DDPG & Price, Hand-crafted & Stock & USA & 1 Day \\ \hline
\cite{wang2019alphastock} & - & Technical Indicator & Stock & China, USA & - \\ \hline
\cite{yu2019model} & Model-based RL & OHLC, Hnad-crafted & Stock & USA & 1 Hour \\ \hline
\cite{shi2019multi} & RRL & Price, Lagged Portfolio & Cryptocurrency & - & - \\ \hline
\cite{benhamou2020bridging} & PG & Price, Hand-crafted & - & - & 1 Day \\ \hline
\cite{zhang2020cost} & PG & OHLC, Lagged Return & Cryptocurrency & - & - \\ \hline
\cite{yang2020deep} & PPO, A2C, DDPG & Price, Hand-crafted & Stock & USA & 1 Day \\ \hline
\cite{lee2020maps} & Multi-agent RL & Technical Indicator & Stock & USA & 1 Day \\ \hline
\cite{sawhney2021quantitative} & DDPG & \tabincell{c}{Financial Text \\ Hand-crafted} & Stock & USA, China & 1 Min \\ \hline
\cite{ye2020reinforcement} & DPG & Financial Text, OHLC & \tabincell{c}{Stock \\ Cryptocurrency} & \tabincell{c}{USA \\ -} & \tabincell{c}{1 Day \\ 30 Min} \\ \hline
\cite{xu2020relation} & RRL & OHLC & \tabincell{c}{Stock \\ Cryptocurrency} & \tabincell{c}{USA \\ -} & 30 Min \\ \hline
\cite{wang2020commission} & REINFORCE,DQN & OHLC, Lagged Portfolio & Stock & USA, China & 1 Day  \\ \hline
\cite{bisi2019risk} & TRPO & \tabincell{c}{Lagged Return \\ Lagged Portfolio} & \tabincell{c}{Artificial \\ Stock Index} & \tabincell{c}{- \\ USA} & \tabincell{c}{- \\ 1 Day} \\ \hline
\cite{wang2021deeptrader} & PG & Technical Indicator & Stock & \tabincell{c}{USA, China \\ Hong Kong} & - \\ \hline 
\cite{huang2021modularized} & Multi-agent RL & Price, Financial Text & Stock & USA  & 1 Day \\ \hline

\end{tabular}
\caption{Summary of RL for Portfolio Management}\label{pm_2}
\end{table}

\subsection{RL in Order Execution}
Different from AT and PM, order execution (OE) is a micro-level QT task, which tries to trade a fixed amount of shares in a given time horizon and minimize the execution cost. In the real financial market, OE is extremely important for institutional traders whose trading volume is large enough to have an obvious impact of the market price.

\citet{nevmyvaka2006reinforcement} proposed the first RL-based method for large-scale order execution. The authors use Q-learning to train the agent with real-world LOB data. With carefully designed state, action and reward function, the Q-learning framework can significantly outperform traditional baselines. \citet{hendricks2014reinforcement} implemented another Q-learning based RL method on South Africa market by extending the popular Almgren-Chriss model with linear price impact. \citet{ning2018double} proposed an RL framework using Double DQN and evaluated its performance on 9 different U.S. stocks. \citet{daberius2019deep} implemented DDQN, PPO and compared their performance with TWAP.

PPO is another widely used RL method for OE. \citet{linend} proposed an end-to-end PPO-based framework. MLP and LSTM are tested as time dependencies accounting network. The authors design a sparse reward function instead of previous implementation shortfall (IS) or a shaped reward function, which leads to state-of-the-art performance on 14 stocks in the U.S. market. \citet{fang2021universal} proposed another PPO-based framework to bridge the gap between the noisy yet imperfect market states and the optimal action sequences for OE. The framework leverages a policy distillation method with an entropy regularization term in the loss function to guide the student agent toward learning similar policy by an oracle teacher with perfect information of the financial market. Moreover, the authors design a normalized reward function to encourage universal learning among different stocks. Extensive experiments on Chinese stock market demonstrate that the proposed method significantly outperforms various baselines with reasonable trading actions. 

We present a summary of existing RL-based order execution publications in Table \ref{oe_2}. Although there are a few successful examples using either Q-learning or PPO on order execution, existing works share a few limitations. First, most of algorithms are only tested on stock data. Their performance on different financial assets (e.g., futures and cryptocurrency) is still unclear. Second, the execution time window (e.g., 1 day) is too long, which makes the task easier. In practice, professional traders usually finish the execution process in much shorter time window (e.g., 10 minute). Third, existing works will fail when the trading volume is huge, because all of them assume there is no obvious market impact, which is impossible for large volume settings. In the real-world, the requirement of institutional investors is to execute large amount of shares in a relatively short time window. There is still a long way to go for researchers to tackle these limitations.

\begin{table}[htbp!]
\newcommand{\tabincell}[2]{\begin{tabular}{@{}#1@{}}#2\end{tabular}}  
\begin{tabular}{cccccc}
\hline
Reference & RL method & Data Source & Asset Type & Market  & Data frequency \\
\hline
\cite{nevmyvaka2006reinforcement} & Q-learning & Price, Hand-crafted & Stock & USA & Millisecond \\ \hline
\cite{hendricks2014reinforcement} & Q-learning & Hand-crafted & Stock & South Africa & 5 Min \\ \hline
\cite{ning2018double} & DQN & LOB & Stock & USA & 1 Second \\ \hline
\cite{daberius2019deep} & DDQN, PPO & Hand-crafted & Artificial & - & - \\ \hline
\cite{linend} & PPO & LOB & Stock & USA & Millisecond \\ \hline
\cite{fang2021universal} & PPO & OHLC, Hand-crafted & Stock & China & 1 Min \\ \hline
\end{tabular}
\caption{Summary of RL for Order Execution}\label{oe_2}
\end{table}

\subsection{RL in Market Making}
Market making refers to trading activities that buy and sell one given asset simultaneously at desired price. The goal of a market maker is to provide liquidity to the market and market profit through the tiny price spread of buy/sell orders. In this subsection, we will discuss existing RL-based methods for market making.  

\citet{chan2001electronic} made the first attempt to apply RL for market making without any assumption of the market. Simulation showed that the RL method converged on optimal strategies successfully on a few controlled environments. \citet{spooner2018market} focused on designing and analyzing temporal-difference (TD) RL methods for market making. The authors firstly build a realistic, data-driven simulator with millisecond LOB data for market making. With an asymmetrically dampened reward function and a linear combination of tile coding as state, both Q-learning and SARSA outperform previous baselines. \citet{lim2018reinforcement} proposed a Q-learning based algorithm with a novel usage of CARA utility as the terminal reward for market making. \citet{gueant2019deep} proposed a model-based actor-critic RL algorithm, which focuses on market making optimization for multiple corporate bonds. \citet{zhongdata} proposed a model-free and off-policy Q-learning algorithm to develop trading strategy implemented with a simple lookup table. The method achieves great performance on event-by-event LOB data confirmed by a professional trading firm. For training robust market making agents, \citet{spooner2020robust} introduced a game-theoretic adaptation of the traditional mathematical market making model. The authors thoroughly investigate the impact of 3 environmental settings with adversarial RL. 

Even though market making is a fundamental task in quantitative trading, research on RL-based market making is still at the early stage. Existing few works simply apply different RL methods on their own data. The summary of order execution publications is in Table \ref{mm_2}. To fully realize the potential of RL for market making, one major obstacle is the lack of high-fidelity micro-level market simulator. At present, there is still no reasonable way to simulate the ubiquitous market impact. This unignorable gap between simulation and real market limits the usage of RL in market making.

\begin{table}[htbp!]
\newcommand{\tabincell}[2]{\begin{tabular}{@{}#1@{}}#2\end{tabular}}  
\begin{tabular}{cccccc}
\hline
Reference & RL method & Data Source & Asset Type & Market  & Data frequency \\
\hline
\cite{chan2001electronic} & SARSA & Hand-crafted & Artificial & - & - \\ \hline
\cite{spooner2018market} & Q-learning, SARSA & Hand-crafted & - & - & Millisecond \\ \hline
\cite{lim2018reinforcement} & Q-learning & Hand-crafted & Artificial & - & - \\ \hline
\cite{gueant2019deep} & Model-based RL & Hand-crafted & Bond & Europe & - \\ \hline
\cite{zhongdata} & Q-learning & LOB & - & - & Event \\ \hline
\cite{spooner2020robust} & SARSA & Hand-crafted & Artificial & - & - \\ \hline

\end{tabular}
\caption{Summary of RL for Market Making}\label{mm_2}
\end{table}

\section{Open Issues and Future Directions}
Even though existing works have demonstrated the success of RL methods on QT tasks, this section will point out a few prospective future research directions. Several critical open issues and potential solutions are also elaborated.



\subsection{Advanced RL techniques on QT}
Most existing works are only straightforward application of classic RL methods on QT tasks. The effectiveness of more advanced RL techniques on financial data is not well-explored. We point out a few promising directions in this subsection.

First, data scarcity is a major challenge on applying RL for QT tasks. Model-based RL can speed up the training process by learning a model of the financial market \cite{yu2019model}. The worst-case (e.g., financial crisis) can be used as a regularizer for maximizing the accumulated reward. Second, the key objective of QT is to balance between maximizing profit and minimizing risk. Multi-objective RL techniques provide a weapon to balance the trade-off between profit and risk. Training diversified trading policies with different risk tolerance is an interesting direction. Third, graph learning \cite{wu2020comprehensive} has shown promising results on modeling the ubiquitous relationship between stocks in supervised learning \cite{feng2019temporal,sawhney2021exploring}. Combing graph learning with RL for modeling the internal relationship between different stocks or financial market is an interesting future direction. Fourth, the severe distribution shift of financial market makes RL-based methods exhibit poor generalization ability in new market condition. Meta-RL and transfer learning techniques can help improve RL-based QT models' generalization performance across different financial assets or market. Fifth, for high risky decision-making tasks such as QT, we need to explain its actions to human traders as a condition for their full acceptance of the algorithm. Hierarchical RL methods decompose the main goal into sub-goals for low-level agents. By learning the optimal subgoals for the low-level agent, the high-level agent forms a representation of the financial market that is interpretable by human traders. Sixth, for QT, learning through directly interacting with the real market is risky and impractical. RL-based QT normally use historical data to learn a policy, which fits in offline RL settings. Offline RL techniques can help to model the distribution shift and risk of financial market while training RL agents.

\subsection{Alternative Data and New QT Settings}
Intuitively, alternative data can provide extra information to learn better representation of the financial market. Economic news \cite{hu2018listening}, frequency of prices \cite{zhang2017stock}, social media \cite{xu2018stock}, financial events \cite{ding2016knowledge} and investment behaviors \cite{chen2019investment} have been applied to improve performance of financial prediction. For RL-based methods, price movement embedding \cite{ye2020reinforcement} and market condition embedding \cite{wang2021deeptrader} are incorporated as extra information to improve an RL agent's performance. However, existing works simply concatenate extra features or embedding from multiple data source as market representations, one interesting forward-looking direction is to utilize multi-modality learning techniques to learn more meaningful representations with both original price and alternative data while training RL agents. Besides alternative data, there are still some important QT settings unexplored by RL researchers. Intraday trading, high-frequency trading and pairs trading are a few examples. Intraday trading tries to capture price fluctuation patterns within the same trading day; high-frequency trading aims at capturing the fleeting micro-level trading opportunities; Pairs trading focus on analyzing the relative trend of two highly correlated assets.

\subsection{Enhance with auto-ML}
Due to the noisy nature of financial data and brittleness of RL methods, the success of RL-based QT models highly replies on carefully designed RL components (e.g., reward function) and proper-tuned hyperparameters (e.g., network architecture). As a result, it is still difficult for people without in-depth knowledge of RL such as economists and professional traders to design profitable RL-based trading strategies. Auto-ML, which tries to design high-quality ML applications automatically, can enhance the development of RL-based QT from three perspectives: (i) For feature engineering, auto-ML can automatically construct, select and collect meaningful features. (ii) For hyperparameter tuning, auto-ML can automatically search for proper hyperparameters such as update rule, learning rate and reward function. (iii) For neural architecture search, auto-ML can automatically search for suitable neural network architectures for training RL agents. With the assistance of auto-ML, RL-based QT models can be more usable for people without in depth knowledge of RL. We believe that it is a promising research direction to facilitate the development of RL-based QT models with auto-ML techniques.    

\subsection{More Realistic Simulation}
High-fidelity simulation is the key foundation of RL methods' success. Although existing works take many practical constraints such as transaction fee \cite{vittori2020dealing}, execution cost \cite{wang2020commission} and slippage \cite{liu2020adaptive} into consideration, current simulation is far from realistic. The ubiquitous market impact, which refers to the effect of one trader's actions to other traders, is ignored. For leading trading firms, their trading volume can account for over 10\% of the total volume with a significant impact of other traders in the market. As a result, simulation with only historical market data is not enough. There are some research efforts focusing on dealing with the market impact. \citet{spooner2018market} tried to take market impact into consideration for MM with event-level data. \citet{byrd2019abides} proposed Aides, an agent-based financial market simulator to model market impact. \citet{vyetrenko2019get} made a survey on current status for market simulation and proposed a series of stylized metrics to test the quality of simulation. It is a very challenging but important research direction to build high-fidelity market simulators. 

\subsection{The Field Needs More Unified and Harder Evaluation}
When a new RL-based QT method is proposed, the authors are expected to compare their methods with SOTA baselines on some financial datasets. At present, the selection of baselines and datasets is seemingly arbitrary, which leads to an inconsistent reporting of revenues. As a result, there is no widely consensus on the general ranking of RL-based methods for QT tasks, which makes it extremely challenging to benchmark new RL algorithms in this field. The question is, how do we solve it? We can borrow some experience from neighbouring ML fields such as computer vision and natural language processing. A suite of standardized evaluation datasets and implementation of SOTA methods could be a good solution to this problem. As for evaluation criteria, although most existing works only evaluate RL algorithms with financial metrics, it is necessary to test RL algorithms on multiple financial assets across different market to evaluate robustness and generalization ability. We also note that split of training, validation and test set in most QT papers is quite random. Since there is significant distribution shift among time in the financial market, it is better to split data on a rolling basis. In addition, it is well-known that the performance of RL methods is very sensitive to hyperparameters such as learning rate. To provide more reliable evaluation of RL methods, authors should spend roughly the same time on tuning hyperparameters for both baselines and their own methods. In practice, some authors make much more effort on tuning their own methods than baselines, which makes the reported revenue not promising. We urge the QT community to conduct more strict evaluation on new proposed methods. With proper datasets, baseline implementation and evaluation scheme, research on RL-based QT could achieve faster development.

\section{Conclusion}
In this article, we provided a comprehensive review of the most notable works on RL-based QT models. We proposed a classification scheme for organizing and clustering existing works, and we highlighted a bunch of influential research prototypes. We also discussed the pros/cons of utilizing RL techniques for QT tasks. In addition, we point out some of the most pressing open problems and promising future directions. Both RL and QT are ongoing hot research topics in the past few decades. There are many newly developing techniques and emerging models each year. We hope that this survey can provide readers with a comprehensive understanding of the key aspects of this field, clarify the most notable advances, and shed some lights on future research.

\bibliographystyle{ACM-Reference-Format}
\bibliography{reference}










\end{document}